\pdfoutput=1

\documentclass[11pt]{article}

\usepackage[]{EMNLP2023}

\usepackage{times}
\usepackage{latexsym}

\usepackage[T1]{fontenc}

\usepackage[utf8]{inputenc}

\usepackage{microtype}

\usepackage{inconsolata}

\usepackage{booktabs}
\usepackage{multirow}
\usepackage{graphicx}

\usepackage{algorithm}
\usepackage{algorithmic}

\usepackage{newfloat}
\usepackage{listings}

\usepackage{subfigure}
\usepackage{amsthm,amsmath,amssymb} 
\usepackage{mathrsfs}
\usepackage{lineno}
\usepackage{xcolor}
\usepackage{color, soul}
\usepackage[inline]{enumitem}
\usepackage{acronym}
\usepackage{enumitem}
\acrodef{LJP}{Legal Judgment Prediction}
\acrodef{MLJP}{Multi-defendant Legal Judgment Prediction}
\acrodef{CMLJ}{Chain of Multi-defendant Legal Judgment}
\acrodef{HRN}{Hierarchical Reasoning Network}
\acrodef{FID}{Fusion-in-Decoder}
\acrodef{Seq2Seq}{Sequence-to-Sequence}

\newcommand{\header}[1]{\vspace{1.5mm}\noindent\textbf{#1}.}

\title{Multi-Defendant Legal Judgment Prediction via Hierarchical Reasoning}

\author{Yougang Lyu\textsuperscript{\rm 1}\thanks{\quad Equal contribution.}\qquad Jitai Hao\textsuperscript{\rm 1}\footnotemark[1]\qquad Zihan Wang\textsuperscript{\rm 1}\qquad  Kai Zhao\textsuperscript{\rm 2}\qquad  Shen Gao\textsuperscript{\rm 1}  \\{\bf Pengjie Ren\textsuperscript{\rm 1}\qquad Zhumin Chen\textsuperscript{\rm 1}\qquad  Fang Wang\textsuperscript{\rm 1}\qquad}  {\bf Zhaochun Ren}\textsuperscript{\rm 3}\thanks{\quad Corresponding author.} 
\\
        \textsuperscript{\rm 1}Shandong University, Qingdao, China 
\\ \textsuperscript{\rm 2}Georgia State University, Atlanta, USA 
\\ \textsuperscript{\rm 3}Leiden University, Leiden, The Netherlands
\\
\{youganglyu, 202215112, 202020630\}@mail.sdu.edu.cn
\\kzhao4@gsu.edu, shengao@sdu.edu.cn, jay.ren@outlook.com
\\ \{chenzhumin, wangfang226\}@sdu.edu.cn, z.ren@liacs.leidenuniv.nl
} 

\begin{document}
\maketitle

\begin{abstract}
Multiple defendants in a criminal fact description generally exhibit complex interactions, and cannot be well handled by existing \ac{LJP} methods which focus on predicting judgment results (e.g., law articles, charges, and terms of penalty) for single-defendant cases. To address this problem, we propose the task of multi-defendant \ac{LJP}, which aims to automatically predict the judgment results for each defendant of multi-defendant cases. Two challenges arise with the task of multi-defendant \ac{LJP}: (1) indistinguishable judgment results among various defendants; and (2) the lack of a real-world dataset for training and evaluation. To tackle the first challenge, we formalize the multi-defendant judgment process as hierarchical reasoning chains and introduce a multi-defendant \ac{LJP} method, named \ac{HRN}, which follows the hierarchical reasoning chains to determine criminal relationships, sentencing circumstances, law articles, charges, and terms of penalty for each defendant. To tackle the second challenge, we collect a real-world multi-defendant \ac{LJP} dataset, namely MultiLJP, to accelerate the relevant research in the future. Extensive experiments on MultiLJP verify the effectiveness of our proposed \ac{HRN}.
\end{abstract}
\section{Introduction}
\begin{figure}[t]
  \centering
  \includegraphics[width=0.95\linewidth]{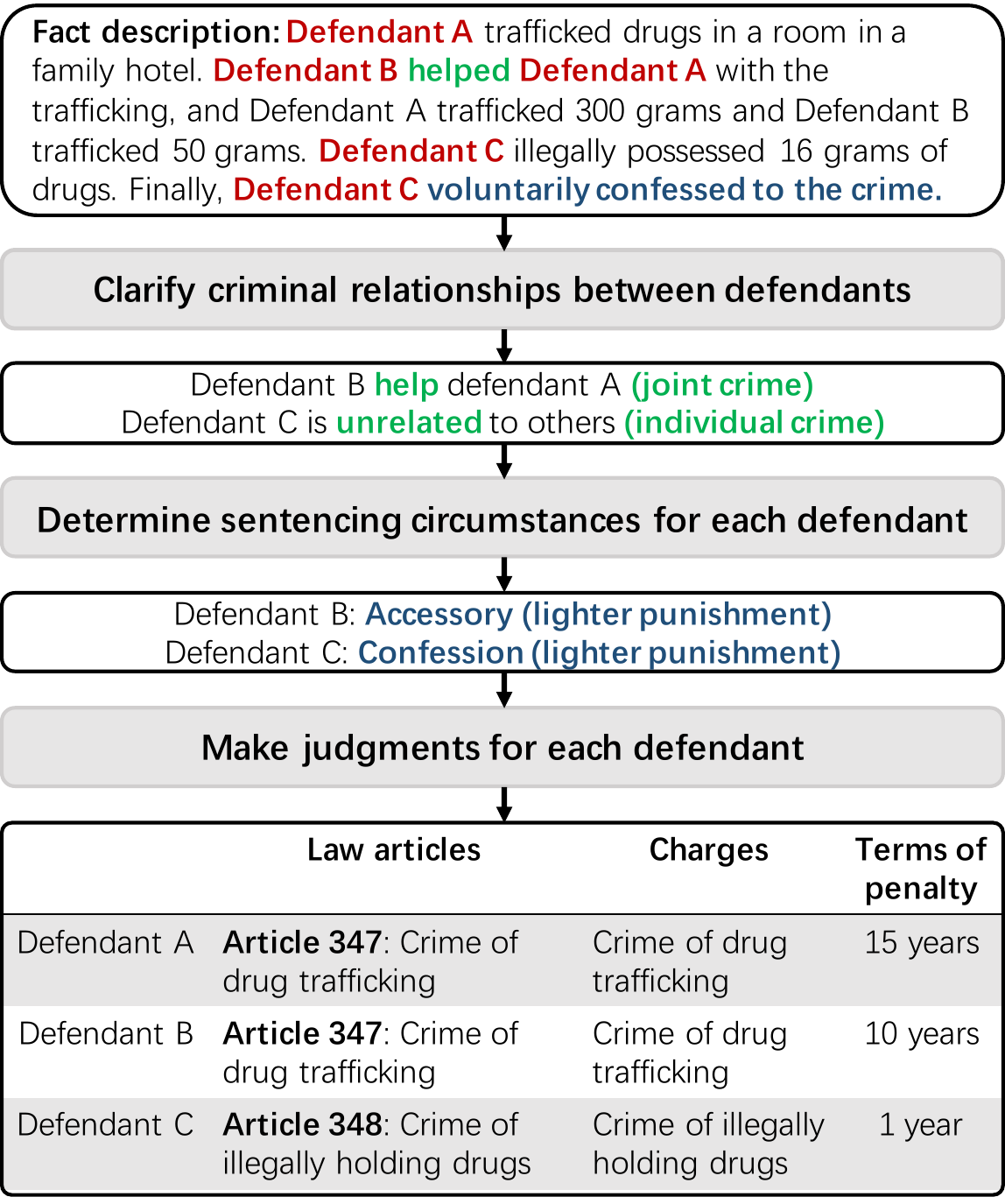}
  \vspace*{-2mm}
  \caption{An illustration of multi-defendant LJP. Generally, a judge needs to reason on the fact description to clarify the complex interactions among different defendants and make accurate judgments for each defendant.
  }
  \label{fig:mljp}
  \vspace*{-3mm}  
\end{figure}
\acf{LJP} aims at predicting judgment results (e.g., law articles, applicable charges, and terms of penalty) based on the fact description of a given case.
Existing \ac{LJP} studies primarily focus on single-defendant cases, where only one defendant is involved~\cite{DBLP:conf/emnlp/LuoFXZZ17,DBLP:conf/emnlp/ZhongGTX0S18, DBLP:conf/ijcai/YangJZL19,DBLP:conf/acl/XuWCPWZ20,DBLP:conf/sigir/DongN21, DBLP:conf/sigir/Yue0JWZACYW21, DBLP:journals/ipm/LyuWRRCLLLS22, DBLP:conf/acl/FengL022,zhang2023contrastive}.

Despite these successful efforts, single-defendant LJP suffers from an inevitable restriction in practice: a large number of fact descriptions with multiple defendants. According to statistics derived from published legal documents sampled from legal case information disclosure, multi-defendant cases constitute a minimum of 30\% of all cases~\cite{DBLP:conf/ccscw/PanLGZX19}. As shown in Figure~\ref{fig:mljp}, the multi-defendant \ac{LJP} task aims at predicting law articles, charges, and terms of penalty for each defendant in multi-defendant cases. 
Since multiple defendants are mentioned in the fact description and exhibit complex interactions, the multi-defendant \ac{LJP} task requires clarifying these interactions and making accurate judgments for each defendant, which is intuitively beyond the reach of single-defendant \ac{LJP} methods. Hence, there is a pressing need to extend \ac{LJP} from single-defendant to multi-defendant scenarios.

However, two main challenges arise with the task of multi-defendant \ac{LJP}:
\begin{itemize}[leftmargin=*,nosep]
\item \header{Indistinguishable judgment results among various defendants} As complicated interactions exist among multiple defendants, fact descriptions of various defendants are usually mixed together. Thus it is difficult to distinguish different judgment results among various defendants so as to make accurate judgments for each defendant.
As shown in Figure~\ref{fig:mljp}, in order to distinguish different judgment results among various defendants, a judge has to clarify criminal relationships among defendants to determine whether defendants share same law articles and charges, and sentencing circumstances affecting terms of penalty for each defendant.
Based on these intermediate reasoning results, the judge determines and verifies the judgment results (law articles, charges, and terms of penalty) for each defendant, following a forward and backward order. The motivation behind forward prediction and backward verification is rooted in the complex nature of legal reasoning, where evidence and conclusions can be interdependent~\cite{DBLP:conf/emnlp/ZhongGTX0S18, DBLP:conf/ijcai/YangJZL19}. Overall, the multi-defendant judgment process requires simulating the judicial logic of human judges and modeling complex reasoning chains.
\item \header{Lack of real-world multi-defendant LJP datasets} Existing datasets for LJP either only have single-defendant cases annotated for multiple LJP subtasks or multi-defendant cases for a single LJP subtask. \citet{DBLP:journals/corr/abs-1807-02478} collect a real-world \ac{LJP} dataset CAIL, which only retains single-defendant cases. \citet{DBLP:conf/ccscw/PanLGZX19} only annotate multi-defendant cases with the charge prediction subtask and ignore information about criminal relationships and sentencing circumstances that can distinguish judgment results of multiple defendants in real scenarios. In order to accelerate the research on multi-defendant \ac{LJP}, we urgently need a real-world multi-defendant \ac{LJP} dataset. 
\end{itemize}

To tackle the first challenge, we formalize the multi-defendant judgment process as hierarchical reasoning chains and propose a method for multi-defendant \ac{LJP}, named \acf{HRN}, which follows the hierarchical reasoning chains to distinguish different judgment results of various defendants.
Specifically, the hierarchical reasoning chains are divided into two levels.
The first-level reasoning chain identifies the relationships between defendants and determines the sentencing circumstances for each defendant.
The second-level reasoning chain predicts and verifies the law articles, charges, and terms of penalty for each defendant, using a forward prediction process and a backward verification process, respectively. Since generative language models have shown great ability to reason~\citep{DBLP:conf/nips/TalmorTCGB20, DBLP:conf/nips/YaoCYJR21,DBLP:journals/corr/abs-2102-02201}, we convert these reasoning chains into \ac{Seq2Seq} generation tasks and apply the mT5~\citep{DBLP:conf/naacl/XueCRKASBR21} to model them.
Furthermore, we adopt \ac{FID}~\citep{DBLP:conf/eacl/IzacardG21} to process multi-defendant fact descriptions with thousands of tokens efficiently.

To tackle the second challenge, we collect a real-world dataset, namely MultiLJP, with 23,717 real-world multi-defendant LJP cases. Eight professional annotators are involved in manually editing law articles, charges, terms of penalty, criminal relationships, and sentencing circumstances for each defendant. In 89.58 percent of these cases, the defendants have different judgment results for at least one of the subtasks of the multi-defendant LJP task. MultiLJP requires accurate distinction of the judgment results for each defendant. This makes MultiLJP different from the existing single-defendant LJP datasets. Our work provides the first benchmark for the multi-defendant LJP task.

Using MultiLJP, we evaluate the effectiveness of \ac{HRN} for multi-defendant \ac{LJP} on various subtasks. The results show that HRN can significantly outperform all the baselines. In summary, our main contributions are:
\begin{itemize}[leftmargin=*,nosep]
    \item We focus on the multi-defendant \ac{LJP} task and formalize the multi-defendant judgment process as hierarchical reasoning chains for the multi-defendant LJP task.
    \item We introduce \ac{HRN}, a novel method that follows the hierarchical reasoning chains to distinguish the judgment results for each defendant in multi-defendant \ac{LJP}.
    \item We present MultiLJP, the first real-world dataset for multi-defendant \ac{LJP}, which facilitates future research in this area\footnote{Our code and MultiLJP dataset are available at \url{https://github.com/CURRENTF/HRN}.}.
    \item We demonstrate the effectiveness of \ac{HRN} on MultiLJP through empirical experiments.
\end{itemize}
\section{Related work}
\label{sec:related_work}

\subsection{Legal judgment prediction}
Legal judgment prediction has been studied in various jurisdictions \citep{DBLP:conf/aaai/ZhongWTZ0S20, DBLP:conf/ijcai/FengL022, katz2017general, DBLP:conf/acl/ChalkidisAA19,sulea2017exploring,DBLP:conf/ranlp/SuleaZVG17, DBLP:conf/acl/MalikSNGGBM20, DBLP:conf/coling/PaulGG20, DBLP:journals/corr/abs-2110-00806}. Early studies on LJP focus on rule-based methods \citep{kort1957predicting,nagel1963applying, segal1984predicting} and machine learning algorithms \citep{DBLP:journals/peerj-cs/AletrasTPL16,sulea2017exploring,DBLP:conf/ranlp/SuleaZVG17,katz2017general}. Recent neural-based approaches jointly predict judgment results (law articles, charges and term of penalty) for single-defendant cases by modeling dependency between LJP subtasks
\citep{DBLP:conf/emnlp/ZhongGTX0S18, DBLP:conf/ijcai/YangJZL19,DBLP:conf/sigir/DongN21, DBLP:journals/corr/abs-2112-06370}, leveraging legal knowledge \citep{DBLP:conf/coling/HuLT0S18, DBLP:conf/aaai/GanKYW21, DBLP:conf/sigir/Yue0JWZACYW21, DBLP:conf/sigir/MaZWLYSZ21, DBLP:journals/ipm/LyuWRRCLLLS22, DBLP:conf/acl/FengL022}, exploiting label information \citep{DBLP:conf/emnlp/LuoFXZZ17, DBLP:conf/sigir/WangFNYZG19,DBLP:conf/acl/XuWCPWZ20, DBLP:conf/cikm/LeZCQ0022, DBLP:conf/coling/LiuDLP022,zhang2023contrastive}, or employing pre-trained language models \citep{DBLP:conf/emnlp/ChalkidisFMAA20, DBLP:conf/naacl/ChalkidisFTAAM21, DBLP:journals/aiopen/XiaoHLTS21}. For multi-defendant cases, MAMD~\cite{DBLP:conf/ccscw/PanLGZX19} utilizes multi-scale attention to distinguish confusing fact descriptions of different defendants for multi-defendant charge prediction.  

However, existing single-defendant \ac{LJP} methods neglect the complex interactions among multiple defendants.  Moreover, compared with the multi-defendant \ac{LJP} method MAMD~\cite{DBLP:conf/ccscw/PanLGZX19}, we follow the human judgment process to model hierarchical reasoning chains to distinguish different judgment results of various defendants and make accurate judgments for each defendant.

\subsection{Multi-step reasoning with language models}
Multi-step reasoning by training or fine-tuning language models to generate intermediate steps has been shown to improve performance \citep{DBLP:conf/naacl/ZaidanEP07,DBLP:conf/nips/TalmorTCGB20, DBLP:conf/nips/YaoCYJR21,DBLP:journals/corr/abs-2102-02201,DBLP:conf/wsdm/Zhang0LLRCMR23, DBLP:journals/corr/abs-2112-08656}. \citet{DBLP:conf/acl/LingYDB17} generate natural language intermediate steps to address math word problems. \citet{DBLP:conf/nips/CamburuRLB18} extend the natural language inference dataset with human-annotated natural language explanations of the entailment relations. \citet{DBLP:conf/acl/RajaniMXS19} generate rationales that explain model predictions for commonsense question-answering tasks \citep{DBLP:conf/naacl/TalmorHLB19}. \citet{DBLP:conf/nips/HendrycksBKABTS21} finetune pre-trained language models to solve competition mathematics problems by generating multi-step solutions. \citet{DBLP:journals/corr/abs-2112-00114} train language models to predict the final outputs of programs by  predicting intermediate computational results. Recently, \citet{DBLP:journals/corr/abs-2201-11903} propose chain of thought prompting, which feed large language models with step-by-step reasoning examples without fine-tuning to improve model performance.

However, these methods are not designed for more realistic legal reasoning applications \citep{DBLP:journals/corr/abs-2212-10403}. In this paper, we aim to use generative language models to capture hierarchical reasoning chains for the multi-defendant \ac{LJP} task.
\begin{table}[t]
\centering \small
\begin{tabular}{@{}lr@{}}
\toprule
MultiLJP  & Number\\ \midrule
\# Training set cases       & 18,968  \\
\# Validation set cases     & 2,379 \\
\# Testing set cases        & 2,370  \\
\# Law articles             & 22    \\
\# Charges                  & 23     \\
\# Terms of penalty          & 11     \\
\# Criminal relationships & 2    \\ 
\# Sentencing circumstances & 8     \\ 
Total defendants & 80,477 \\
Average defendants & 3.39 \\
Law articles per defendant & 1.06 \\
Charges per defendant & 1.06 \\
Terms of penalty per defendant & 1 \\
Criminal relationships per defendant & 0.88 \\
Sentencing circumstances per defendant & 1.19 \\
Average length & 3,040.76    \\
\bottomrule
\end{tabular}
\vspace*{-2mm}
\caption{Statistics of the MultiLJP.}
\vspace*{-3mm}
\label{tab:dataset}
\end{table}
\section{Dataset}
In this section, we describe the construction process and analyze various aspects of MultiLJP to provide a deeper understanding of the dataset.
\subsection{Dataset construction}
To the best of our knowledge, existing \ac{LJP} datasets only focus on single-defendant cases or charge prediction for multi-defendant cases.
Thus, we construct a Multi-defendant Legal Judgment Prediction (MultiLJP) dataset from the published legal documents in China Judgements Online\footnote{\url{https://wenshu.court.gov.cn/}}. 
Instead of extracting labels using regular expressions as in existing works~\cite{DBLP:journals/corr/abs-1807-02478,DBLP:conf/ccscw/PanLGZX19}, we hire eight professional annotators to manually produce law articles, charges, terms of penalty, criminal relationships, and sentencing circumstances for each defendant in multi-defendant cases. 
The annotators are native Chinese speakers who have passed China’s Unified Qualification Exam for Legal Professionals. 
All data is evaluated by two annotators repeatedly to eliminate bias. Since second-instance cases and retrial cases are too complicated, we only retain first-instance cases. 
Besides, we anonymize sensitive information (e.g., name, location, etc.) for multi-defendant cases to avoid potential risks of structured social biases~\cite{DBLP:journals/sigmod/PitouraTFFPAW17} and protect personal privacy. 
After preprocessing and manual annotation, the MultiLJP consists of 23,717 multi-defendant cases. The statistics information of dataset MultiLJP can be found in Table~\ref{tab:dataset}.

\subsection{Dataset analysis} 
\header{Analysis of the number of defendants}
MultiLJP only contains multi-defendant cases. The number of defendants per case is distributed as follows: 49.40 percent of cases have two defendants, 21.41 percent have three defendants, 11.22 percent have four defendants, and 17.97 percent have more than four defendants. The MultiLJP dataset has 80,477 defendants in total. On average, each multi-defendant case has 3.39 defendants.

\header{Analysis of multi-defendant judgment results}
In 89.58 percent of these cases, the defendants have different judgment results for at least one of the subtasks of the multi-defendant LJP task. Specifically, 18.91 percent of cases apply different law articles to different defendants, 26.80 percent of cases impose different charges on different defendants, and 88.54 percent of cases assign different terms of penalty to different defendants.

\header{Analysis of criminal relationships and sentencing circumstances}
Based on the gold labels of criminal relationships and sentencing circumstances, ideally, a judge can distinguish between 69.73 percent of defendants with different judgment results (law articles, charges, and terms of penalty). 
Specifically, based on the criminal relationship, a judge can distinguish 70.28 percent of defendants with different law articles and 72.50 percent of defendants with different charges; based on the sentencing circumstances, a judge can distinguish 96.28 percent of defendants with different terms of penalty.

\begin{figure*}[htbp]
  \centering
\includegraphics[width=1.0\textwidth]{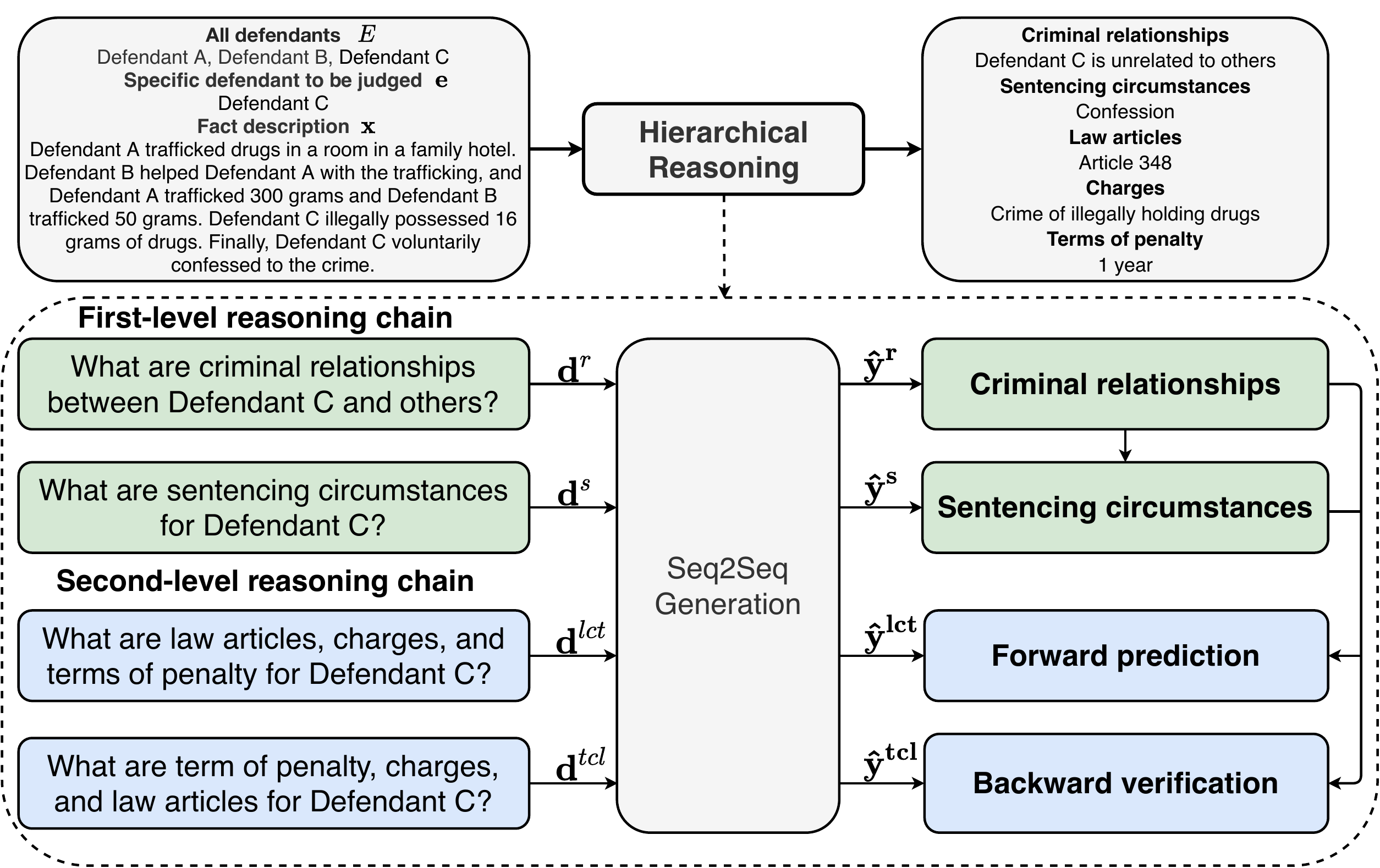}
\vspace*{-7mm}
\caption{Overview of our proposed HRN. HRN leverages Sequence-to-Sequence (Seq2Seq) generation framework to follow hierarchical reasoning chains to generate prediction results.} 
\label{fig:2}
\vspace*{-4mm}
\end{figure*}
\section{Method}
\label{sec:method}
In this section, we describe the \ac{HRN} method. 
First, we formulate our research problem.
Then, we introduce the \acf{Seq2Seq} generation framework for hierarchical reasoning. 
Next, we introduce hierarchical reasoning chains of multi-defendant in detail. Finally, a training process with Fusion-in-Decoder for \ac{HRN} is  explained.
\subsection{Problem formulation}
\label{ssec:problem_formulatin}
We first formulate the multi-defendant \ac{LJP} task. The fact description of a multi-defendant case can be seen as a word sequence $\mathbf{x} =\{w_{1},w_{2},...,w_{n}\}$, where n represents the number of words. Each multi-defendant case has a set of defendant names $E=\{\mathbf{e}_{1},\mathbf{e}_{2},...,\mathbf{e}_{|E|}\}$, where each name is a sequence of words $\mathbf{e}=\{w_{1},w_{2},..,w_{|\mathbf{e}|}\}$. Given the fact description $\mathbf{x}$ and the defendant name $\mathbf{e}$ of a multi-defendant case, the multi-defendant task aims to predict the judgment results of multiple applicable law articles, multiple charges, and a term of penalty.  
The law article prediction and the charge prediction subtasks are multi-label classification problems, and the term of penalty prediction subtask is a multi-class classification problem.

We introduce the criminal relationship and the sentencing circumstance as intermediate tasks to model the hierarchical reasoning chains for multi-defendant \ac{LJP} and improve the prediction of the main judgment results. Criminal relationships refer to the relationships between defendants, specifically whether one defendant assisted other co-defendants during the commission of the crime. Sentencing circumstances refer to specific behaviors (such as confession and recidivist) or factors (like accessory and blind) that can influence the severity or leniency of a penalty\footnote{Details of criminal relationships and sentencing circumstances definitions together with
their explanations can be found in Appendix~\ref{appendix:cr} and~\ref{appendix:sc}.}.
These two tasks are also multi-label classification problems. We denote the labels of law articles, charges, terms of penalty, criminal relationships, and sentencing circumstances as word sequences $\mathbf{y}^{l},\mathbf{y}^{c},\mathbf{y}^{t},\mathbf{y}^{r}$ and $\mathbf{y}^{s}$ respectively in this paper.

\subsection{Sequence-to-sequence generation}
\label{ssec:overview}
From the perspective of \acf{Seq2Seq} generation, each task can be modeled as finding an optimal label sequence $\mathbf{y}$ that maximizes the conditional probability based on the fact description, a specific defendant name and a specific task description $p(\mathbf{y}|\mathbf{x},\mathbf{e},\mathbf{d})$, which is calculated as follows:
\begin{equation} 
\label{eq:1}
p(\mathbf{y}|\mathbf{x},\mathbf{e},\mathbf{d}) =\prod_{i=1}^{m} p(y_{i}|y_{1},y_{2},...,y_{m-1},\mathbf{x},\mathbf{e},\mathbf{d}),
\end{equation}
where $m$ denotes the length of the label sequence, and the specific task description $\mathbf{d}$ is a semantic prompt that allows \ac{Seq2Seq} generation models to execute the desired task. To accomplish the \ac{Seq2Seq} generation tasks, we apply the \ac{Seq2Seq} language model mT5~\cite{DBLP:conf/naacl/XueCRKASBR21} to generate label sequences as follows:
\begin{equation} 
\label{eq:2}
\mathbf{\hat{y}}=DEC(ENC(\mathbf{x},\mathbf{e},\mathbf{d})),
\end{equation}
where $ENC$ refers to the encoder of the language model, $DEC$ denotes the decoder of the language model, and $\mathbf{\hat{y}}$ is prediction results composed of words. We use special [SEP] tokens to separate the different information to form the input of the encoder.

\subsection{Hierarchical reasoning chains}
\label{ssec:hierarchical_learning}
To distinguish different judgment results among various defendants, our method \ac{HRN} follows hierarchical reasoning chains to determine each defendant's criminal relationships, sentencing circumstances, law articles, charges, and terms of penalty. As shown in Figure~\ref{fig:2}, the hierarchical reasoning chains consist of two levels:

\textbf{The first-level reasoning is for intermediate tasks.} The first-level reasoning chain is to first identify relationships between defendants based on the fact description, the names of all defendants and the criminal relationship prediction task description $d^{r}$ as follows:
\begin{equation} 
\label{eq:3}
\mathbf{\hat{y}^{r}}=DEC(ENC(\mathbf{x},E,\mathbf{d}^{r})).
\end{equation}
Then, we determine sentencing circumstances for the defendant $\mathbf{e}$ based on the fact description, name of the defendant $\mathbf{e}$, prediction results of criminal relationships and the sentencing circumstance prediction task description $\mathbf{d}^{s}$, that is:
\begin{equation} 
\label{eq:4}
\mathbf{\hat{y}^{s}}=DEC(ENC(\mathbf{x},\mathbf{e},\mathbf{\hat{y}}^{r},\mathbf{d}^{s}).
\end{equation}

\textbf{The second-level reasoning is for judgment prediction tasks.} The second-level reasoning chain consists of a forward prediction process and a backward verification process. The forward prediction process is to predict law articles, charges, and terms of penalty (in that order) based on the fact description, the name of defendant $\mathbf{e}$, first-level reasoning results, and the forward prediction task description $\mathbf{d}^{lct}$ as follows:
\begin{equation} 
\label{eq:5}
\mathbf{\hat{y}^{lct}}=DEC(ENC(\mathbf{x},\mathbf{e},\mathbf{\hat{y}}^{r},\mathbf{\hat{y}}^{s},\mathbf{d}^{lct})).
\end{equation}
Then, the backward verification process is to verify these judgment results in reverse order based on the fact description, the name of defendant $\mathbf{e}$, first-level reasoning results and the backward verification task description $\mathbf{d}^{tcl}$, that is:
\begin{equation} 
\label{eq:6}
\mathbf{\hat{y}^{tcl}}=DEC(ENC(\mathbf{x},\mathbf{e},\mathbf{\hat{y}}^{r},\mathbf{\hat{y}}^{s},\mathbf{d}^{tcl})).
\end{equation}

\subsection{Training with fusion-in-decoder}
\label{ssec:fusion}
To handle multi-defendant fact descriptions whose average length exceeds the length limit of the encoder, we adopt \acf{FID} \cite{DBLP:conf/eacl/IzacardG21} to encode multiple paragraphs split from a fact description. We first split the fact description $\mathbf{x}$ into $K$ paragraphs containing $M$ words. Then, we combine multiple paragraph representations from the encoder, the decoder generates prediction results by attending to multiple paragraph representations as follows:
\begin{equation} 
\label{eq:7}
\mathbf{\hat{y}}=DEC(\mathbf{h}_{1},\mathbf{h}_{2},...,\mathbf{h}_{K}),
\end{equation}
where $\mathbf{h}_{i}$ denotes the representation of the $i$-th paragraph of the fact description $\mathbf{x}$. Since all tasks are formulated as sequence-to-sequence generation tasks, we follow \citet{DBLP:journals/jmlr/RaffelSRLNMZLL20} to train the model by standard maximum likelihood and calculate the cross-entropy loss for each task. The overall loss function is formally computed as:
\begin{equation} 
\mathcal{L}= \lambda_{r}\mathcal{L}_{r}+\lambda_{s}\mathcal{L}_{s}+\lambda_{lct}\mathcal{L}_{lct}+\lambda_{tcl}\mathcal{L}_{tcl},
\end{equation}
where hyperparameters $\lambda$ determine the trade-off between all subtask losses.
$\mathcal{L}_{r}$, $\mathcal{L}_{s}$, $\mathcal{L}_{lct}$ and $\mathcal{L}_{tcl}$ denote the cross-entropy losses of the criminal relationship prediction task, the sentencing circumstance prediction task, the forward prediction process and the backward verification process, respectively. 
At test time, we apply greedy decoding to generate forward and backward prediction results. Finally, the chain with the highest confidence is chosen for the final prediction.
\begin{table*}[htbp]
\centering \small
\begin{tabular}{lcccccccccccc}
\toprule
\multirow{2}{*}{Method} & \multicolumn{4}{c}{Law Articles} & \multicolumn{4}{c}{Charges} & \multicolumn{4}{c}{Term of Penalty} \\ \cmidrule(r){ 2 - 5 } \cmidrule(r) { 6 - 9 } \cmidrule{ 10 - 13 }  
                          & Acc.   & MP    & MR    & F1    & Acc.   & MP   & MR   & F1   & Acc.     & MP     & MR     & F1     \\ \midrule
TopJudge                & 69.32       & 35.60     & 39.13     & 36.93      & 64.42      & 24.96     & 35.28     & 28.34     & 28.36         & 23.16      & 22.25       & 22.00      \\
MPBFN                  & 72.47        & 34.73      & 34.22      & 34.35      & 65.59       & 32.79     & 33.20     & 31.59      & 28.32         & 21.59       & 20.91       & 20.70      \\
LADAN    & 54.57       & 38.09      & 22.40     & 26.64     &  46.62     & 20.68     & 32.42     & 24.74     & 27.05         & 24.05       & 23.43       & 23.16      \\
NeurJudge & 65.21       & 41.72      & 36.96      & 38.15      & 59.51      & 34.19     & 25.36     & 27.55    & 30.06         & 27.56      & 25.63       & 25.95      \\ 
\midrule
BERT    & 51.38       & 34.19     & 29.68      & 30.70    & 44.80      & 36.80     & 20.10     & 25.14     & 29.60         & 23.95       & 22.68      & 21.55      \\
mT5     & 87.49      & 74.28      & 53.65  & 58.84    & 81.52      & 63.33     & 49.94            & 52.86       & 33.66       & 39.13    & 24.23  & 23.04  \\
Lawformer & 75.50      & 36.18      & 35.33  & 34.00    & 65.94      & 38.97     & 29.12            & 32.76       & 32.37       & 22.66    & 20.68  & 18.30  \\
\midrule
MAMD & -      & -      & -      & -     & 58.73      & 33.00     & 34.15     & 31.60    & -         & -      & -       & -     \\ 
\midrule
\textbf{HRN}          & \textbf{91.46}\rlap{$^{\ast}$}              & \textbf{69.87}\rlap{$^{\ast}$}     & \textbf{70.95}\rlap{$^{\ast}$}      & \textbf{69.20}\rlap{$^{\ast}$}      & \textbf{89.54}\rlap{$^{\ast}$}       &  \textbf{71.80}\rlap{$^{\ast}$}     & \textbf{71.83}\rlap{$^{\ast}$}     & \textbf{70.70}\rlap{$^{\ast}$}     & \textbf{42.74}\rlap{$^{\ast}$}         & \textbf{41.33}\rlap{$^{\ast}$}       & \textbf{40.20}\rlap{$^{\ast}$}       & \textbf{40.62}\rlap{$^{\ast}$} \\
w/ gold  & 92.26\rlap{$^{\ast}$}      & 75.67\rlap{$^{\ast}$}      & 70.22\rlap{$^{\ast}$}  & 71.58\rlap{$^{\ast}$}    & 90.60\rlap{$^{\ast}$}      & 78.46\rlap{$^{\ast}$}     & 75.38\rlap{$^{\ast}$}            & 76.27\rlap{$^{\ast}$}       & 44.44\rlap{$^{\ast}$}       & 47.13\rlap{$^{\ast}$}    & 42.31\rlap{$^{\ast}$}  & 43.27\rlap{$^{\ast}$}

\\

\bottomrule
\end{tabular}
\vspace*{-2mm}
\caption{Judgment prediction results on MultiLJP. Significant improvements over the best baseline are marked with $\ast$ (t-test, $p < 0.05$).}
\vspace*{-2mm}
\label{tab:my-table}
\end{table*}

\section{Experiments}
\label{sec:experimental_setup}
\subsection{Research questions}
We aim to answer the following research questions with our experiments:
(RQ1) How does our proposed method, \ac{HRN}, perform on multi-defendant LJP cases?
(RQ2) How do the different levels of reasoning chains affect the performances of \ac{HRN} on multi-defendant \ac{LJP}?

\subsection{Baselines}
To verify the effectiveness of our method \ac{HRN} on multi-defendant \ac{LJP}, we compare it with a variety of methods, which can be summarized in the following three groups:
\begin{itemize}[leftmargin=*,nosep]
\item \textbf{Single-defendant LJP methods,} including \textbf{Topjudge}~\citep{DBLP:conf/emnlp/ZhongGTX0S18}, which is a topological dependency learning framework for single-defendant LJP and formalizes the explicit dependencies over subtasks as a directed acyclic graph; \textbf{MPBFN}~\citep{DBLP:conf/ijcai/YangJZL19}, which is a single-defendant \ac{LJP} method and utilizes forward and backward dependencies among multiple \ac{LJP} subtasks; \textbf{LADAN}~\citep{DBLP:conf/acl/XuWCPWZ20}, which is a graph neural network based method that automatically captures subtle differences among confusing law articles; \textbf{NeurJudge}~\citep{DBLP:conf/sigir/Yue0JWZACYW21}, which utilizes the results of intermediate subtasks to separate the fact statement into different circumstances and exploits them to make the predictions of other subtasks.
\item \textbf{Pre-trained language models,} including 
\textbf{BERT}~\cite{DBLP:journals/taslp/CuiCLQY21}, which is a Transformer-based method that is pre-trained on Chinese Wikipedia documents; \textbf{mT5}~\cite{DBLP:conf/naacl/XueCRKASBR21}, which is a multilingual model pre-trained by converting several language tasks into “text-to-text” tasks and pre-trained on Chinese datasets; \textbf{Lawformer}~\cite{DBLP:journals/corr/abs-2105-03887}, which is a Transformer-based method that is pre-trained on large-scale Chinese legal long case documents. 
\item \textbf{Multi-defendant charge prediction method,} including \textbf{MAMD}~\cite{DBLP:conf/ccscw/PanLGZX19}, which is a multi-defendant charge prediction method that leverages multi-scale attention to recognize fact descriptions for different defendants.
\end{itemize}

We adapt single-defendant LJP methods to multi-defendant LJP by concatenating a defendant’s name and a fact description as input and training models to predict judgment results. However, we exclude some state-of-the-art single-defendant approaches unsuitable for multi-defendant settings. Few-shot~\cite{DBLP:conf/coling/HuLT0S18}, EPM~\cite{DBLP:conf/acl/FengL022}, and CEEN~\cite{DBLP:journals/ipm/LyuWRRCLLLS22} annotate extra attributes for single-defendant datasets, not easily transferable to MultiLJP. Also, CTM~\cite{DBLP:conf/coling/LiuDLP022} and CL4LJP~\cite{zhang2023contrastive} design specific sampling strategies for contrastive learning of single-defendant cases, hard to generalize to multi-defendant cases.

\subsection{Implementation details}
To accommodate the length of multi-defendant fact descriptions, we set the maximum fact length to 2304. Due to the constraints of the model input, BERT's input length is limited to 512. For training, we employed the AdamW~\cite{DBLP:conf/iclr/LoshchilovH19} optimizer and used a linear learning rate schedule with warmup. The warmup ratio was set to 0.01, and the maximum learning rate was set to $1 \cdot 10^{-3}.$ We set the batch size as 128 and adopt the gradient accumulation strategy. All models are trained for a maximum of 20 epochs. The model that performs best on the validation set is selected. For the hyperparameters, $\lambda$, in the loss function, the best setting is \{0.6, 0.8, 1,4, 1.2\} for \{$\lambda_{r}$, $\lambda_{s}$, $\lambda_{lct}$, $\lambda_{tcl}$\}.
Additionally, we set the number of paragraphs $K$, the number of words per paragraph $M$, and the output length to 3, 768, and 64, respectively. For performance evaluation, we employ four metrics: accuracy (Acc.), Macro-Precision (MP), Macro-Recall (MR), and Macro-F1 (F1). All experiments are conducted on one RTX3090.
\section{Experimental results and analysis}
\label{sec:results}
In this section, we first conduct multi-defendant legal judgment predictions and ablation studies to answer the research questions listed in the Section. 5.1. In addition, we also conducted a case study to intuitively evaluate the importance of hierarchical reasoning.

\begin{table*}[htbp]
\centering \small
\begin{tabular}{lcccccccccccc}
\toprule
\multirow{2}{*}{Method} & \multicolumn{4}{c}{Law Articles} & \multicolumn{4}{c}{Charges} & \multicolumn{4}{c}{Term of Penalty} \\ \cmidrule(r){ 2 - 5 } \cmidrule(r) { 6 - 9 } \cmidrule{ 10 - 13 }  
                          & Acc.   & MP    & MR    & F1    & Acc.   & MP   & MR   & F1   & Acc.     & MP     & MR     & F1     \\ \midrule
HRN                         & 91.46              & 69.87     & 70.95      & 69.20      & 89.54       &  71.80     & 71.83     & 70.70     & 42.74         & 41.33       & 40.20       & 40.62       \\\midrule
w/o CR                     & 89.14      & 73.71      & 67.85  & 68.25           & 86.84     & 72.81     & 68.61      & 68.95         & 37.83       & 38.20   & 30.30    & 30.69   
\\

w/o SC        & 88.58      & 73.64      & 61.51   & 65.51         & 85.98     & 77.03    & 64.82     & 68.98        &  32.66      & 37.37       & 25.43   & 24.56      \\\midrule
w/o FP                             & 89.05      & 73.61     & 65.61            & 68.17     & 85.97    & 73.24     & 66.09         & 68.29       &  35.83      & 34.35  & 30.47 & 30.00   \\
w/o BV                      & 90.91      & 75.60      & 67.58     &  68.50     & 83.45     & 69.34     & 60.86   & 63.73         & 37.41       & 36.90       & 28.42     & 28.88   \\ 
w/o all     & 87.49      & 74.28      & 53.65  & 58.84    & 81.52      & 63.33     & 49.94            & 52.86      & 33.66       & 39.13    & 24.23  & 23.04  \\

\bottomrule
\end{tabular}
\vspace*{-2mm}
\caption{Ablation studies on MultiLJP}
\label{tab:ablation}
\vspace*{-3mm}
\end{table*}

\subsection{Multi-defendant judgment results (RQ1)}
\label{ssec:multi_judgment_prediction_results}
Table~\ref{tab:my-table} shows the evaluation results on the multi-defendant \ac{LJP} subtasks. Generally, \ac{HRN} achieves the best performance in terms of all metrics for all multi-defendant \ac{LJP} subtasks. Based on the results, we have three main observations:
\begin{itemize}[leftmargin=*,nosep]
\item Compared with state-of-art single-defendant \ac{LJP} methods, e.g., Topjudge, MPBFN, LADAN, and NeurJudge, our method \ac{HRN} consider hierarchical reasoning chains and thus achieve significant improvements. Since the single-defendant methods do not consider criminal relationships and sentencing circumstances,  they can not distinguish different judgment results among various defendants well. It indicates the importance of following the hierarchical reasoning chains to predict criminal relationships and sentencing circumstances for multi-defendant \ac{LJP}.
\item As Table~\ref{tab:my-table} shows, our method HRN achieves considerable performance improvements on all subtasks of multi-defendant LJP compared to pre-trained models BERT and Lawformer. This shows that modeling the hierarchical reasoning chains during the fine-tuning stage is crucial. Moreover, HRN, which combines mT5 and the hierarchical reasoning chains, significantly outperforms mT5, indicating that the language knowledge from pre-training and the information of the hierarchical reasoning chains are complementary to each other.
 \item Compared to MAMD designed for multi-defendant charge prediction, our method \ac{HRN} outperforms MAMD on charge prediction task. This shows the effectiveness and robustness of modeling reasoning chains in real-world application scenarios.
 \item We employ gold annotations of criminal relationships and sentencing circumstances, rather than predicted ones, to enhance the performance of \ac{HRN}. Our findings indicate a considerable improvement in results when relying on gold annotations. This observation underscores the potential for notable enhancement in multi-defendant LJP by improving the first-level reasoning.
\end{itemize}

\subsection{Ablation studies (RQ2)}
\label{ssec:ablation_study}
To analyze the effect of the different levels of reasoning chains in \ac{HRN}, we conduct an ablation study. 
Table~\ref{tab:ablation} shows the results on MultiLJP with five settings: \begin{enumerate*}[label=(\roman*)]
\item w/o CR: HRN without predicting criminal relationships.
\item w/o SC: HRN without predicting sentencing circumstances.
\item w/o FP: HRN without the forward prediction process and predicts law articles, charges, and terms of penalty in reverse order.
\item w/o BV: HRN without the backward verification process and predicts law articles, charges, and terms of penalty in order. 
\item w/o all: HRN degrades to mT5 with multi-task settings by removing all reasoning chains.
\end{enumerate*}

\begin{figure}[t]
  \centering
  \includegraphics[width=1\linewidth]{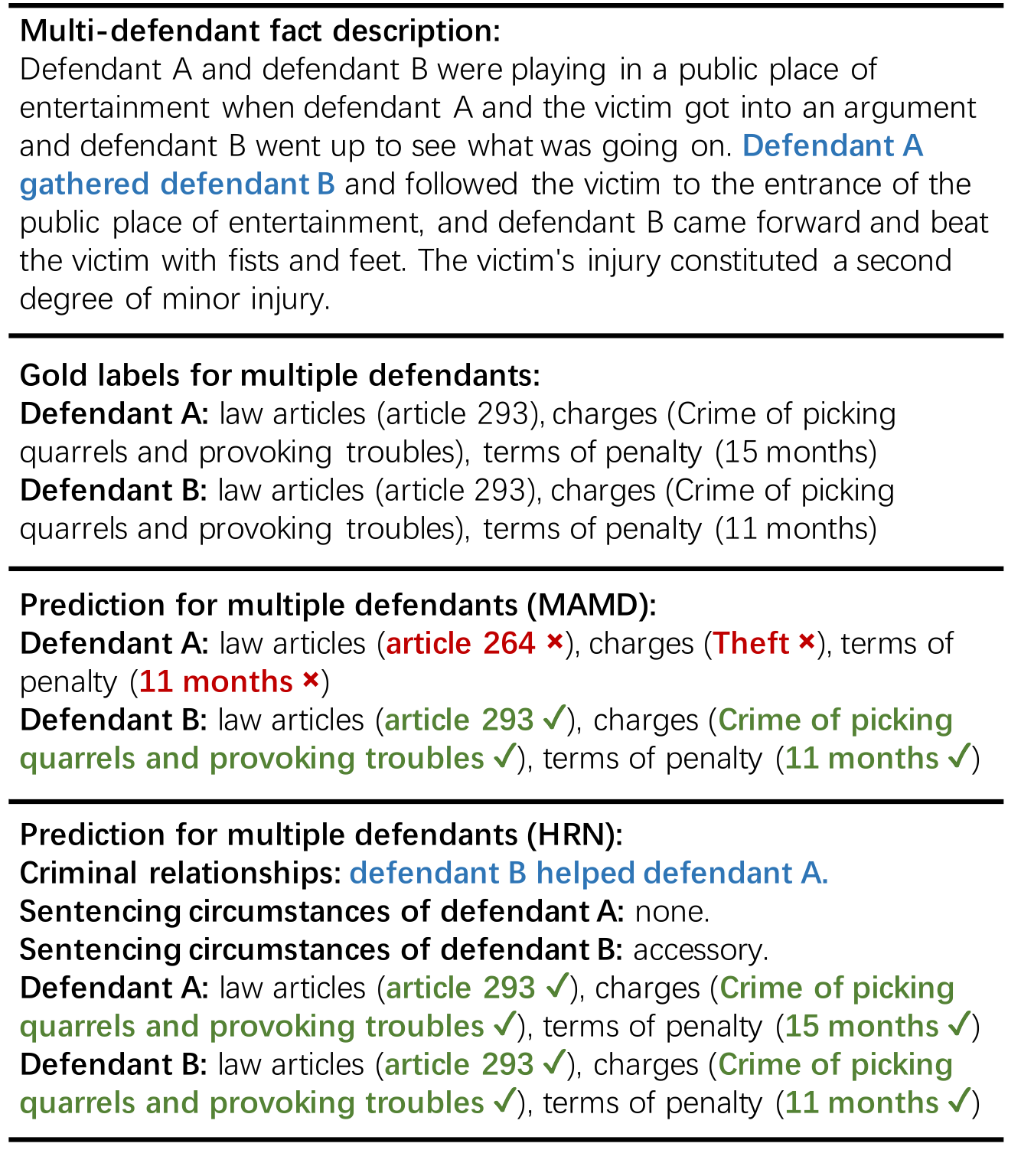}
  \vspace*{-6mm}
  \caption{Case study for intuitive comparisons. Red and green represent incorrect and correct
judgment results, respectively. Blue denotes descriptions of criminal relationships.} 
  \label{fig:case_study}
  \vspace*{-3mm}  
\end{figure}

Table~\ref{tab:ablation} shows that all levels of reasoning chains help HRN as removing any of them decreases performance: 
\begin{itemize}[leftmargin=*,nosep]
\item \header{Removing the first-level reasoning chain} We observe that both criminal relationships and sentencing circumstances decrease the performance of HRN when we remove the first-level reasoning chains. Specifically, removing criminal relationships (CR) negatively impacts performance, especially on law articles and charges prediction, which means criminal relationships are helpful for distinguishing law articles and charges; removing sentencing circumstances (SC) negatively impacts performance, especially on terms of penalty, which means sentencing circumstances are helpful for distinguishing terms of penalty.
\item \header{Removing the second-level reasoning chain} We observe that the model without the second-level forward prediction process (FP) or the second-level backward verification process (BV) faces a huge performance degradation in multi-defendant \ac{LJP}. As a result, although the model still performs the first-level reasoning, the absence of modeling forward or backward dependencies between LJP subtasks leads to poor LJP performances.
\item \header{Removing all reasoning chains} When removing all reasoning chains from HRN, there is a substantial drop in the performances of multi-defendant \ac{LJP}. Experimental results prove that hierarchical reasoning chains  can be critical for multi-defendant \ac{LJP}.
\end{itemize}

\subsection{Case study}
\label{ssec:case_study}
We also conduct a case study to show how multi-defendant reasoning chains help the model distinguish judgment results of different defendants and make correct predictions. Figure~\ref{fig:case_study} shows prediction results for two defendants, where red and green represent incorrect and correct predictions, respectively. Defendant B helped A beat the victim, but the fact description does not show a direct crime against the victim. Without determining criminal relationships and sentencing circumstances, MAMD focuses on the tailing behavior around A and misclassifies A’s law articles, charges, and terms of penalty as article 264, theft, and 11 months. In contrast, by following reasoning chains to determine criminal relationships, sentencing circumstances, law articles, charges, and terms of penalty, HRN distinguishes different judgment results between defendants.
\section{Conclusions}
\label{sec:conclusion}
In this paper, we studied the legal judgment prediction problem for multi-defendant cases. We proposed the task of multi-defendant \ac{LJP} to promote \ac{LJP} systems from single-defendant to multi-defendant. To distinguish confusing judgment results of different defendants, we proposed a \acf{HRN} to determine criminal relationships, sentencing circumstances, law articles, charges and terms of penalty for each defendant. As there is no benchmark dataset for multi-defendant \ac{LJP}, we have collected a real-world dataset MultiLJP. We conducted extensive experiments on the MultiLJP dataset. Experimental results have verified the effectiveness of our proposed method and \ac{HRN} outperforms all baselines.

\section*{Limitations}
Although our work distinguishes the judgment results of multiple defendants by hierarchical reasoning, in real life, there exist many confusing charge pairs, such as (the crime of intentional injury, and the crime of intentional homicide). The fact descriptions of these confusing charge pairs are very similar, which makes it difficult for the multi-defendant \ac{LJP} model to distinguish between confusing charge pairs. We leave this challenge for future work.

\section*{Ethics Statement}
Since multi-defendant legal judgment prediction is an emerging but sensitive technology, we would like to discuss ethical concerns of our work. Our proposed method \ac{HRN} is a preliminary multi-defendant work and aims to assist legal professionals instead of replacing them. In addition, multi-defendant cases contain personal privacy information. To avoid potential risks of structured social biases \cite{DBLP:journals/sigmod/PitouraTFFPAW17, DBLP:conf/aaai/LyuLYRRZYR23} and protect personal privacy, we have anonymized sensitive information (e.g., name, location, etc.) for multi-defendant cases in MultiLJP dataset.

\section*{Acknowledgments}
This work was supported by 
the National Key R\&D Program of China with grant No.2022YFC3303004, the Natural Science Foundation of China (62272274, 61902219, 61972234, 62102234, 62202271, 62072279, T2293773, 72371145),
the Natural Science Foundation of Shandong Province (ZR2021QF129),
the Key Scientific and Technological Innovation Program of Shandong Province (2019JZZY010129). All content represents the opinion of the authors, which is not necessarily shared or endorsed by their respective employers and/or sponsors.

\bibliography{custom}

\begin{thebibliography}{56}
\expandafter\ifx\csname natexlab\endcsname\relax\def\natexlab#1{#1}\fi

\bibitem[{Aletras et~al.(2016)Aletras, Tsarapatsanis, Preotiuc{-}Pietro, and Lampos}]{DBLP:journals/peerj-cs/AletrasTPL16}
Nikolaos Aletras, Dimitrios Tsarapatsanis, Daniel Preotiuc{-}Pietro, and Vasileios Lampos. 2016.
\newblock \href {https://doi.org/10.7717/peerj-cs.93} {Predicting judicial decisions of the european court of human rights: a natural language processing perspective}.
\newblock \emph{PeerJ Comput. Sci.}, 2:e93.

\bibitem[{Camburu et~al.(2018)Camburu, Rockt{\"{a}}schel, Lukasiewicz, and Blunsom}]{DBLP:conf/nips/CamburuRLB18}
Oana{-}Maria Camburu, Tim Rockt{\"{a}}schel, Thomas Lukasiewicz, and Phil Blunsom. 2018.
\newblock \href {https://proceedings.neurips.cc/paper/2018/hash/4c7a167bb329bd92580a99ce422d6fa6-Abstract.html} {e-snli: Natural language inference with natural language explanations}.
\newblock In \emph{Proceedings of NeurIPS}, pages 9560--9572.

\bibitem[{Chalkidis et~al.(2019)Chalkidis, Androutsopoulos, and Aletras}]{DBLP:conf/acl/ChalkidisAA19}
Ilias Chalkidis, Ion Androutsopoulos, and Nikolaos Aletras. 2019.
\newblock \href {https://doi.org/10.18653/v1/p19-1424} {Neural legal judgment prediction in english}.
\newblock In \emph{Proceedings of ACL}, pages 4317--4323.

\bibitem[{Chalkidis et~al.(2020)Chalkidis, Fergadiotis, Malakasiotis, Aletras, and Androutsopoulos}]{DBLP:conf/emnlp/ChalkidisFMAA20}
Ilias Chalkidis, Manos Fergadiotis, Prodromos Malakasiotis, Nikolaos Aletras, and Ion Androutsopoulos. 2020.
\newblock \href {https://doi.org/10.18653/v1/2020.findings-emnlp.261} {{LEGAL-BERT:} "preparing the muppets for court'"}.
\newblock In \emph{Findings of EMNLP}, pages 2898--2904.

\bibitem[{Chalkidis et~al.(2021)Chalkidis, Fergadiotis, Tsarapatsanis, Aletras, Androutsopoulos, and Malakasiotis}]{DBLP:conf/naacl/ChalkidisFTAAM21}
Ilias Chalkidis, Manos Fergadiotis, Dimitrios Tsarapatsanis, Nikolaos Aletras, Ion Androutsopoulos, and Prodromos Malakasiotis. 2021.
\newblock \href {https://doi.org/10.18653/v1/2021.naacl-main.22} {Paragraph-level rationale extraction through regularization: {A} case study on european court of human rights cases}.
\newblock In \emph{Proceedings of NAACL}, pages 226--241.

\bibitem[{Cui et~al.(2021)Cui, Che, Liu, Qin, and Yang}]{DBLP:journals/taslp/CuiCLQY21}
Yiming Cui, Wanxiang Che, Ting Liu, Bing Qin, and Ziqing Yang. 2021.
\newblock \href {https://doi.org/10.1109/TASLP.2021.3124365} {Pre-training with whole word masking for chinese {BERT}}.
\newblock \emph{{IEEE} {ACM} Trans. Audio Speech Lang. Process.}, 29:3504--3514.

\bibitem[{Dong and Niu(2021)}]{DBLP:conf/sigir/DongN21}
Qian Dong and Shuzi Niu. 2021.
\newblock \href {https://doi.org/10.1145/3404835.3462931} {Legal judgment prediction via relational learning}.
\newblock In \emph{Proceedings of SIGIR}, pages 983--992.

\bibitem[{Feng et~al.(2022{\natexlab{a}})Feng, Li, and Ng}]{DBLP:conf/ijcai/FengL022}
Yi~Feng, Chuanyi Li, and Vincent Ng. 2022{\natexlab{a}}.
\newblock \href {https://doi.org/10.24963/ijcai.2022/765} {Legal judgment prediction: {A} survey of the state of the art}.
\newblock In \emph{Proceedings of IJCAI}, pages 5461--5469.

\bibitem[{Feng et~al.(2022{\natexlab{b}})Feng, Li, and Ng}]{DBLP:conf/acl/FengL022}
Yi~Feng, Chuanyi Li, and Vincent Ng. 2022{\natexlab{b}}.
\newblock \href {https://doi.org/10.18653/v1/2022.acl-long.48} {Legal judgment prediction via event extraction with constraints}.
\newblock In \emph{Proceedings of ACL}, pages 648--664.

\bibitem[{Gan et~al.(2021)Gan, Kuang, Yang, and Wu}]{DBLP:conf/aaai/GanKYW21}
Leilei Gan, Kun Kuang, Yi~Yang, and Fei Wu. 2021.
\newblock \href {https://ojs.aaai.org/index.php/AAAI/article/view/17522} {Judgment prediction via injecting legal knowledge into neural networks}.
\newblock In \emph{Proceedings of AAAI}, pages 12866--12874.

\bibitem[{Gu et~al.(2021)Gu, Mishra, and Clark}]{DBLP:journals/corr/abs-2112-08656}
Yuling Gu, Bhavana~Dalvi Mishra, and Peter Clark. 2021.
\newblock \href {https://arxiv.org/abs/2112.08656} {{DREAM:} uncovering mental models behind language models}.
\newblock \emph{CoRR}, abs/2112.08656.

\bibitem[{Hase and Bansal(2021)}]{DBLP:journals/corr/abs-2102-02201}
Peter Hase and Mohit Bansal. 2021.
\newblock \href {https://arxiv.org/abs/2102.02201} {When can models learn from explanations? {A} formal framework for understanding the roles of explanation data}.
\newblock \emph{CoRR}, abs/2102.02201.

\bibitem[{Hendrycks et~al.(2021)Hendrycks, Burns, Kadavath, Arora, Basart, Tang, Song, and Steinhardt}]{DBLP:conf/nips/HendrycksBKABTS21}
Dan Hendrycks, Collin Burns, Saurav Kadavath, Akul Arora, Steven Basart, Eric Tang, Dawn Song, and Jacob Steinhardt. 2021.
\newblock \href {https://datasets-benchmarks-proceedings.neurips.cc/paper/2021/hash/be83ab3ecd0db773eb2dc1b0a17836a1-Abstract-round2.html} {Measuring mathematical problem solving with the {MATH} dataset}.
\newblock In \emph{Proceedings of NeurIPS}.

\bibitem[{Hu et~al.(2018)Hu, Li, Tu, Liu, and Sun}]{DBLP:conf/coling/HuLT0S18}
Zikun Hu, Xiang Li, Cunchao Tu, Zhiyuan Liu, and Maosong Sun. 2018.
\newblock \href {https://aclanthology.org/C18-1041/} {Few-shot charge prediction with discriminative legal attributes}.
\newblock In \emph{Proceedings of COLING}, pages 487--498.

\bibitem[{Huang and Chang(2022)}]{DBLP:journals/corr/abs-2212-10403}
Jie Huang and Kevin~Chen{-}Chuan Chang. 2022.
\newblock \href {https://doi.org/10.48550/arXiv.2212.10403} {Towards reasoning in large language models: {A} survey}.
\newblock \emph{CoRR}, abs/2212.10403.

\bibitem[{Huang et~al.(2021)Huang, Shen, Li, Ge, and Luo}]{DBLP:journals/corr/abs-2112-06370}
Yunyun Huang, Xiaoyu Shen, Chuanyi Li, Jidong Ge, and Bin Luo. 2021.
\newblock \href {https://arxiv.org/abs/2112.06370} {Dependency learning for legal judgment prediction with a unified text-to-text transformer}.
\newblock \emph{CoRR}, abs/2112.06370.

\bibitem[{Izacard and Grave(2021)}]{DBLP:conf/eacl/IzacardG21}
Gautier Izacard and Edouard Grave. 2021.
\newblock \href {https://doi.org/10.18653/v1/2021.eacl-main.74} {Leveraging passage retrieval with generative models for open domain question answering}.
\newblock In \emph{Proceedings of EACL}, pages 874--880.

\bibitem[{Katz et~al.(2017)Katz, Bommarito, and Blackman}]{katz2017general}
Daniel~Martin Katz, Michael~J Bommarito, and Josh Blackman. 2017.
\newblock A general approach for predicting the behavior of the supreme court of the united states.
\newblock \emph{PloS one}, 12(4):e0174698.

\bibitem[{Kort(1957)}]{kort1957predicting}
Fred Kort. 1957.
\newblock Predicting supreme court decisions mathematically: A quantitative analysis of the" right to counsel" cases.
\newblock \emph{The American Political Science Review}, 51(1):1--12.

\bibitem[{Le et~al.(2022)Le, Zhao, Chen, Quan, He, and Li}]{DBLP:conf/cikm/LeZCQ0022}
Yuquan Le, Yuming Zhao, Meng Chen, Zhe Quan, Xiaodong He, and Kenli Li. 2022.
\newblock \href {https://doi.org/10.1145/3511808.3557379} {Legal charge prediction via bilinear attention network}.
\newblock In \emph{Proceedings of CIKM}, pages 1024--1033.

\bibitem[{Ling et~al.(2017)Ling, Yogatama, Dyer, and Blunsom}]{DBLP:conf/acl/LingYDB17}
Wang Ling, Dani Yogatama, Chris Dyer, and Phil Blunsom. 2017.
\newblock \href {https://doi.org/10.18653/v1/P17-1015} {Program induction by rationale generation: Learning to solve and explain algebraic word problems}.
\newblock In \emph{Proceedings of ACL}, pages 158--167.

\bibitem[{Liu et~al.(2022)Liu, Du, Li, Pan, and Ming}]{DBLP:conf/coling/LiuDLP022}
Dugang Liu, Weihao Du, Lei Li, Weike Pan, and Zhong Ming. 2022.
\newblock \href {https://aclanthology.org/2022.coling-1.235} {Augmenting legal judgment prediction with contrastive case relations}.
\newblock In \emph{Proceedings of COLING}, pages 2658--2667.

\bibitem[{Loshchilov and Hutter(2019)}]{DBLP:conf/iclr/LoshchilovH19}
Ilya Loshchilov and Frank Hutter. 2019.
\newblock \href {https://openreview.net/forum?id=Bkg6RiCqY7} {Decoupled weight decay regularization}.
\newblock In \emph{Proceedings of ICLR}.

\bibitem[{Luo et~al.(2017)Luo, Feng, Xu, Zhang, and Zhao}]{DBLP:conf/emnlp/LuoFXZZ17}
Bingfeng Luo, Yansong Feng, Jianbo Xu, Xiang Zhang, and Dongyan Zhao. 2017.
\newblock \href {https://doi.org/10.18653/v1/d17-1289} {Learning to predict charges for criminal cases with legal basis}.
\newblock In \emph{Proceedings of EMNLP}, pages 2727--2736.

\bibitem[{Lyu et~al.(2023)Lyu, Li, Yang, de~Rijke, Ren, Zhao, Yin, and Ren}]{DBLP:conf/aaai/LyuLYRRZYR23}
Yougang Lyu, Piji Li, Yechang Yang, Maarten de~Rijke, Pengjie Ren, Yukun Zhao, Dawei Yin, and Zhaochun Ren. 2023.
\newblock \href {https://doi.org/10.1609/aaai.v37i11.26567} {Feature-level debiased natural language understanding}.
\newblock In \emph{Proceedings of AAAI}, pages 13353--13361.

\bibitem[{Lyu et~al.(2022)Lyu, Wang, Ren, Ren, Chen, Liu, Li, Li, and Song}]{DBLP:journals/ipm/LyuWRRCLLLS22}
Yougang Lyu, Zihan Wang, Zhaochun Ren, Pengjie Ren, Zhumin Chen, Xiaozhong Liu, Yujun Li, Hongsong Li, and Hongye Song. 2022.
\newblock \href {https://doi.org/10.1016/j.ipm.2021.102780} {Improving legal judgment prediction through reinforced criminal element extraction}.
\newblock \emph{Inf. Process. Manag.}, 59(1):102780.

\bibitem[{Ma et~al.(2021)Ma, Zhang, Wang, Liu, Ye, Sun, and Zhang}]{DBLP:conf/sigir/MaZWLYSZ21}
Luyao Ma, Yating Zhang, Tianyi Wang, Xiaozhong Liu, Wei Ye, Changlong Sun, and Shikun Zhang. 2021.
\newblock \href {https://doi.org/10.1145/3404835.3462945} {Legal judgment prediction with multi-stage case representation learning in the real court setting}.
\newblock In \emph{Proceedings of SIGIR}, pages 993--1002.

\bibitem[{Malik et~al.(2021)Malik, Sanjay, Nigam, Ghosh, Guha, Bhattacharya, and Modi}]{DBLP:conf/acl/MalikSNGGBM20}
Vijit Malik, Rishabh Sanjay, Shubham~Kumar Nigam, Kripabandhu Ghosh, Shouvik~Kumar Guha, Arnab Bhattacharya, and Ashutosh Modi. 2021.
\newblock \href {https://doi.org/10.18653/v1/2021.acl-long.313} {{ILDC} for {CJPE:} indian legal documents corpus for court judgment prediction and explanation}.
\newblock In \emph{Proceedings of ACL}, pages 4046--4062.

\bibitem[{Nagel(1963)}]{nagel1963applying}
Stuart~S Nagel. 1963.
\newblock Applying correlation analysis to case prediction.
\newblock \emph{Tex. L. Rev.}, 42:1006.

\bibitem[{Niklaus et~al.(2021)Niklaus, Chalkidis, and St{\"{u}}rmer}]{DBLP:journals/corr/abs-2110-00806}
Joel Niklaus, Ilias Chalkidis, and Matthias St{\"{u}}rmer. 2021.
\newblock \href {https://arxiv.org/abs/2110.00806} {Swiss-judgment-prediction: {A} multilingual legal judgment prediction benchmark}.
\newblock \emph{CoRR}, abs/2110.00806.

\bibitem[{Nye et~al.(2021)Nye, Andreassen, Gur{-}Ari, Michalewski, Austin, Bieber, Dohan, Lewkowycz, Bosma, Luan, Sutton, and Odena}]{DBLP:journals/corr/abs-2112-00114}
Maxwell~I. Nye, Anders~Johan Andreassen, Guy Gur{-}Ari, Henryk Michalewski, Jacob Austin, David Bieber, David Dohan, Aitor Lewkowycz, Maarten Bosma, David Luan, Charles Sutton, and Augustus Odena. 2021.
\newblock \href {https://arxiv.org/abs/2112.00114} {Show your work: Scratchpads for intermediate computation with language models}.
\newblock \emph{CoRR}, abs/2112.00114.

\bibitem[{Pan et~al.(2019)Pan, Lu, Gu, Zhang, and Xu}]{DBLP:conf/ccscw/PanLGZX19}
Sicheng Pan, Tun Lu, Ning Gu, Huajuan Zhang, and Chunlin Xu. 2019.
\newblock \href {https://doi.org/10.1007/978-981-15-1377-0\_59} {Charge prediction for multi-defendant cases with multi-scale attention}.
\newblock In \emph{Proceedings of ChineseCSCW}, volume 1042 of \emph{Communications in Computer and Information Science}, pages 766--777.

\bibitem[{Paul et~al.(2020)Paul, Goyal, and Ghosh}]{DBLP:conf/coling/PaulGG20}
Shounak Paul, Pawan Goyal, and Saptarshi Ghosh. 2020.
\newblock \href {https://doi.org/10.18653/v1/2020.coling-main.88} {Automatic charge identification from facts: {A} few sentence-level charge annotations is all you need}.
\newblock In \emph{Proceedings of COLING}, pages 1011--1022.

\bibitem[{Pitoura et~al.(2017)Pitoura, Tsaparas, Flouris, Fundulaki, Papadakos, Abiteboul, and Weikum}]{DBLP:journals/sigmod/PitouraTFFPAW17}
Evaggelia Pitoura, Panayiotis Tsaparas, Giorgos Flouris, Irini Fundulaki, Panagiotis Papadakos, Serge Abiteboul, and Gerhard Weikum. 2017.
\newblock \href {https://doi.org/10.1145/3186549.3186553} {On measuring bias in online information}.
\newblock \emph{{SIGMOD} Rec.}, 46(4):16--21.

\bibitem[{Raffel et~al.(2020)Raffel, Shazeer, Roberts, Lee, Narang, Matena, Zhou, Li, and Liu}]{DBLP:journals/jmlr/RaffelSRLNMZLL20}
Colin Raffel, Noam Shazeer, Adam Roberts, Katherine Lee, Sharan Narang, Michael Matena, Yanqi Zhou, Wei Li, and Peter~J. Liu. 2020.
\newblock \href {http://jmlr.org/papers/v21/20-074.html} {Exploring the limits of transfer learning with a unified text-to-text transformer}.
\newblock \emph{J. Mach. Learn. Res.}, 21:140:1--140:67.

\bibitem[{Rajani et~al.(2019)Rajani, McCann, Xiong, and Socher}]{DBLP:conf/acl/RajaniMXS19}
Nazneen~Fatema Rajani, Bryan McCann, Caiming Xiong, and Richard Socher. 2019.
\newblock \href {https://doi.org/10.18653/v1/p19-1487} {Explain yourself! leveraging language models for commonsense reasoning}.
\newblock In \emph{Proceedings of ACL}, pages 4932--4942.

\bibitem[{Segal(1984)}]{segal1984predicting}
Jeffrey~A Segal. 1984.
\newblock Predicting supreme court cases probabilistically: The search and seizure cases, 1962-1981.
\newblock \emph{American Political Science Review}, 78(4):891--900.

\bibitem[{Sulea et~al.(2017{\natexlab{a}})Sulea, Zampieri, Malmasi, Vela, Dinu, and van Genabith}]{sulea2017exploring}
Octavia{-}Maria Sulea, Marcos Zampieri, Shervin Malmasi, Mihaela Vela, Liviu~P. Dinu, and Josef van Genabith. 2017{\natexlab{a}}.
\newblock Exploring the use of text classification in the legal domain.
\newblock In \emph{Proceedings of ICAIL}, volume 2143.

\bibitem[{Sulea et~al.(2017{\natexlab{b}})Sulea, Zampieri, Vela, and van Genabith}]{DBLP:conf/ranlp/SuleaZVG17}
Octavia{-}Maria Sulea, Marcos Zampieri, Mihaela Vela, and Josef van Genabith. 2017{\natexlab{b}}.
\newblock \href {https://doi.org/10.26615/978-954-452-049-6\_092} {Predicting the law area and decisions of french supreme court cases}.
\newblock In \emph{Proceedings of RANLP}, pages 716--722.

\bibitem[{Talmor et~al.(2019)Talmor, Herzig, Lourie, and Berant}]{DBLP:conf/naacl/TalmorHLB19}
Alon Talmor, Jonathan Herzig, Nicholas Lourie, and Jonathan Berant. 2019.
\newblock \href {https://doi.org/10.18653/v1/n19-1421} {Commonsenseqa: {A} question answering challenge targeting commonsense knowledge}.
\newblock In \emph{Proceedings of NAACL}, pages 4149--4158.

\bibitem[{Talmor et~al.(2020)Talmor, Tafjord, Clark, Goldberg, and Berant}]{DBLP:conf/nips/TalmorTCGB20}
Alon Talmor, Oyvind Tafjord, Peter Clark, Yoav Goldberg, and Jonathan Berant. 2020.
\newblock \href {https://proceedings.neurips.cc/paper/2020/hash/e992111e4ab9985366e806733383bd8c-Abstract.html} {Leap-of-thought: Teaching pre-trained models to systematically reason over implicit knowledge}.
\newblock In \emph{Proceedings of NeurIPS}.

\bibitem[{Wang et~al.(2019)Wang, Fan, Niu, Yang, Zhang, and Guo}]{DBLP:conf/sigir/WangFNYZG19}
Pengfei Wang, Yu~Fan, Shuzi Niu, Ze~Yang, Yongfeng Zhang, and Jiafeng Guo. 2019.
\newblock \href {https://doi.org/10.1145/3331184.3331223} {Hierarchical matching network for crime classification}.
\newblock In \emph{Proceedings of SIGIR}, pages 325--334.

\bibitem[{Wei et~al.(2022)Wei, Wang, Schuurmans, Bosma, Chi, Le, and Zhou}]{DBLP:journals/corr/abs-2201-11903}
Jason Wei, Xuezhi Wang, Dale Schuurmans, Maarten Bosma, Ed~H. Chi, Quoc Le, and Denny Zhou. 2022.
\newblock \href {https://arxiv.org/abs/2201.11903} {Chain of thought prompting elicits reasoning in large language models}.
\newblock \emph{CoRR}, abs/2201.11903.

\bibitem[{Xiao et~al.(2021{\natexlab{a}})Xiao, Hu, Liu, Tu, and Sun}]{DBLP:journals/aiopen/XiaoHLTS21}
Chaojun Xiao, Xueyu Hu, Zhiyuan Liu, Cunchao Tu, and Maosong Sun. 2021{\natexlab{a}}.
\newblock \href {https://doi.org/10.1016/j.aiopen.2021.06.003} {Lawformer: {A} pre-trained language model for chinese legal long documents}.
\newblock \emph{{AI} Open}, 2:79--84.

\bibitem[{Xiao et~al.(2021{\natexlab{b}})Xiao, Hu, Liu, Tu, and Sun}]{DBLP:journals/corr/abs-2105-03887}
Chaojun Xiao, Xueyu Hu, Zhiyuan Liu, Cunchao Tu, and Maosong Sun. 2021{\natexlab{b}}.
\newblock Lawformer: {A} pre-trained language model for chinese legal long documents.
\newblock \emph{CoRR}, abs/2105.03887.

\bibitem[{Xiao et~al.(2018)Xiao, Zhong, Guo, Tu, Liu, Sun, Feng, Han, Hu, Wang, and Xu}]{DBLP:journals/corr/abs-1807-02478}
Chaojun Xiao, Haoxi Zhong, Zhipeng Guo, Cunchao Tu, Zhiyuan Liu, Maosong Sun, Yansong Feng, Xianpei Han, Zhen Hu, Heng Wang, and Jianfeng Xu. 2018.
\newblock \href {http://arxiv.org/abs/1807.02478} {{CAIL2018:} {A} large-scale legal dataset for judgment prediction}.
\newblock \emph{CoRR}, abs/1807.02478.

\bibitem[{Xu et~al.(2020)Xu, Wang, Chen, Pan, Wang, and Zhao}]{DBLP:conf/acl/XuWCPWZ20}
Nuo Xu, Pinghui Wang, Long Chen, Li~Pan, Xiaoyan Wang, and Junzhou Zhao. 2020.
\newblock \href {https://doi.org/10.18653/v1/2020.acl-main.280} {Distinguish confusing law articles for legal judgment prediction}.
\newblock In \emph{Proceedings of ACL}, pages 3086--3095.

\bibitem[{Xue et~al.(2021)Xue, Constant, Roberts, Kale, Al{-}Rfou, Siddhant, Barua, and Raffel}]{DBLP:conf/naacl/XueCRKASBR21}
Linting Xue, Noah Constant, Adam Roberts, Mihir Kale, Rami Al{-}Rfou, Aditya Siddhant, Aditya Barua, and Colin Raffel. 2021.
\newblock \href {https://doi.org/10.18653/v1/2021.naacl-main.41} {mt5: {A} massively multilingual pre-trained text-to-text transformer}.
\newblock In \emph{Proceedings of NAACL}, pages 483--498.

\bibitem[{Yang et~al.(2019)Yang, Jia, Zhou, and Luo}]{DBLP:conf/ijcai/YangJZL19}
Wenmian Yang, Weijia Jia, Xiaojie Zhou, and Yutao Luo. 2019.
\newblock \href {https://doi.org/10.24963/ijcai.2019/567} {Legal judgment prediction via multi-perspective bi-feedback network}.
\newblock In \emph{Proceedings of IJCAI}, pages 4085--4091.

\bibitem[{Yao et~al.(2021)Yao, Chen, Ye, Jin, and Ren}]{DBLP:conf/nips/YaoCYJR21}
Huihan Yao, Ying Chen, Qinyuan Ye, Xisen Jin, and Xiang Ren. 2021.
\newblock \href {https://proceedings.neurips.cc/paper/2021/hash/4b26dc4663ccf960c8538d595d0a1d3a-Abstract.html} {Refining language models with compositional explanations}.
\newblock In \emph{Proceedings of NeurIPS}, pages 8954--8967.

\bibitem[{Yue et~al.(2021)Yue, Liu, Jin, Wu, Zhang, An, Cheng, Yin, and Wu}]{DBLP:conf/sigir/Yue0JWZACYW21}
Linan Yue, Qi~Liu, Binbin Jin, Han Wu, Kai Zhang, Yanqing An, Mingyue Cheng, Biao Yin, and Dayong Wu. 2021.
\newblock \href {https://doi.org/10.1145/3404835.3462826} {Neurjudge: {A} circumstance-aware neural framework for legal judgment prediction}.
\newblock In \emph{Proceedings of SIGIR}, pages 973--982.

\bibitem[{Zaidan et~al.(2007)Zaidan, Eisner, and Piatko}]{DBLP:conf/naacl/ZaidanEP07}
Omar Zaidan, Jason Eisner, and Christine~D. Piatko. 2007.
\newblock \href {https://aclanthology.org/N07-1033/} {Using "annotator rationales" to improve machine learning for text categorization}.
\newblock In \emph{Proceedings of NAACL}, pages 260--267.

\bibitem[{Zhang et~al.(2023{\natexlab{a}})Zhang, Dou, Zhu, and Wen}]{zhang2023contrastive}
Han Zhang, Zhicheng Dou, Yutao Zhu, and Ji-Rong Wen. 2023{\natexlab{a}}.
\newblock Contrastive learning for legal judgment prediction.
\newblock \emph{ACM Transactions on Information Systems}.

\bibitem[{Zhang et~al.(2023{\natexlab{b}})Zhang, Xin, Li, Liu, Ren, Chen, Ma, and Ren}]{DBLP:conf/wsdm/Zhang0LLRCMR23}
Xiaoyu Zhang, Xin Xin, Dongdong Li, Wenxuan Liu, Pengjie Ren, Zhumin Chen, Jun Ma, and Zhaochun Ren. 2023{\natexlab{b}}.
\newblock \href {https://doi.org/10.1145/3539597.3570426} {Variational reasoning over incomplete knowledge graphs for conversational recommendation}.
\newblock In \emph{Proceedings of WSDM}, pages 231--239. {ACM}.

\bibitem[{Zhong et~al.(2018)Zhong, Guo, Tu, Xiao, Liu, and Sun}]{DBLP:conf/emnlp/ZhongGTX0S18}
Haoxi Zhong, Zhipeng Guo, Cunchao Tu, Chaojun Xiao, Zhiyuan Liu, and Maosong Sun. 2018.
\newblock \href {https://doi.org/10.18653/v1/d18-1390} {Legal judgment prediction via topological learning}.
\newblock In \emph{Proceedings of EMNLP}, pages 3540--3549. Association for Computational Linguistics.

\bibitem[{Zhong et~al.(2020)Zhong, Wang, Tu, Zhang, Liu, and Sun}]{DBLP:conf/aaai/ZhongWTZ0S20}
Haoxi Zhong, Yuzhong Wang, Cunchao Tu, Tianyang Zhang, Zhiyuan Liu, and Maosong Sun. 2020.
\newblock \href {https://ojs.aaai.org/index.php/AAAI/article/view/5479} {Iteratively questioning and answering for interpretable legal judgment prediction}.
\newblock In \emph{Proceedings of AAAI}, pages 1250--1257.

\end{thebibliography}
\bibliographystyle{acl_natbib}

\appendix
\section*{Appendix}
\section{Criminal Relationship}
\label{appendix:cr}
Criminal relationships refer to the relationships between defendants, specifically whether one defendant assisted other co-defendants during the commission of the crime. 
\begin{itemize}[leftmargin=*,nosep]
\item \header{None} One defendant did not help other defendants to conduct criminal behaviors.
\item \header{Assistance} One defendant helped other defendants to conduct criminal behaviors.
\end{itemize}
\section{Sentencing Circumstance}
\label{appendix:sc}
Sentencing circumstances refer to specific behaviors (such as confession and recidivist) or factors (like accessory and blind) that can influence the severity or leniency of a penalty.
\begin{itemize}[leftmargin=*,nosep]
\item \header{Old People} A person who has reached the age of 75 and conducts the crime is liable to a lighter punishment.
\item \header{Deaf-mute or Blind} A deaf-mute or blind person who conducts a crime is liable to a lighter punishment.
\item \header{Accessory} Accessory to the crime is liable to a lighter punishment.
\item \header{Attempted Crime} A person who has attempted to conduct a crime is liable to a lighter punishment.
\item \header{Surrender} A person who has surrendered to justice is liable to a lighter punishment.
\item \header{Confession} A person who has confessed to offence is liable to a lighter punishment.
\item \header{Metitorious Serviceoffset} A person who has metitorious serviceoffset is liable to a lighter punishment.
\item \header{Recidivist} Recidivist is liable to a heavier punishment.
\end{itemize}
\end{document}